\documentclass[runningheads]{llncs}
\usepackage[T1]{fontenc}
\usepackage{booktabs}
\usepackage{multirow} 
\usepackage{graphicx} 
\begin{document}
\title{Enhancing Human-Computer Interaction in Chest X-ray Analysis using Vision and Language Model with Eye Gaze Patterns}
\author{Yunsoo Kim\and
Jinge Wu\and
Yusuf Abdulle\and
Yue Gao\and
Honghan Wu}
\authorrunning{Y. Kim et al.}
\institute{University College London \\
\email{yunsoo.kim.23@ucl.ac.uk}
\vspace{\baselineskip}
}

% \institute{Anonymous Organization \\
% \email{\{******\}@****.****}}
%
\maketitle              % typeset the header of the contribution
\begin{abstract}

Recent advancements in Computer Assisted Diagnosis have shown promising performance in medical imaging tasks, particularly in chest X-ray analysis. However, the interaction between these models and radiologists has been primarily limited to input images. This work proposes a novel approach to enhance human-computer interaction in chest X-ray analysis using Vision-Language Models (VLMs) enhanced with radiologists' attention by incorporating eye gaze data alongside textual prompts. Our approach leverages heatmaps generated from eye gaze data, overlaying them onto medical images to highlight areas of intense radiologist's focus during chest X-ray evaluation. We evaluate this methodology in tasks such as visual question answering, chest X-ray report automation, error detection, and differential diagnosis. Our results demonstrate the inclusion of eye gaze information significantly enhances the accuracy of chest X-ray analysis. Also, the impact of eye gaze on fine-tuning was confirmed as it outperformed other medical VLMs in all tasks except visual question answering. This work marks the potential of leveraging both the VLM's capabilities and the radiologist's domain knowledge to improve the capabilities of AI models in medical imaging, paving a novel way for Computer Assisted Diagnosis with a human-centred AI.

% At least one model showed improvement in the task with eye gaze pattern, and all the baseline models performed better with eye gaze information in differential diagnosis. 
% This multimodal approach fosters a more collaborative diagnostic tool, leveraging both the VLM's capabilities and the radiologist's domain knowledge. 

\keywords{Vision Language Model  \and Eye Gaze \and Chest X-ray}
\end{abstract}
\section{Introduction}
Recent AI breakthroughs have facilitated the development of advanced diagnostic tools such as Computer-Aided Diagnosis (CAD). These systems leverage machine learning and deep learning models to tackle medical challenges \cite{hwang2021deep,cadradiographyreview,cadCOVIDimprovement}. While CAD demonstrated promising results in improving diagnostic accuracy, concerns remain about its reliability and effectiveness as a standalone tool in clinical settings. 

One solution is leveraging human-computer interaction. Studies show integrating human expertise into CAD enhanced accuracy and reliability. This approach outperformed both radiologists and AI models making the decision alone in diagnostic accuracy \cite{calisto2022hitlbreast,hitlperformance}. However, the current AI models used with human-computer interaction for medical image analysis are dominantly limited to analyzing images only, thereby restricting the usage of these models in clinical settings.

A recent breakthrough in Vision-Language Models (VLMs) with Large Language Models (LLMs) extends CAD applications with human-computer interaction to complex multimodal data. This significant advancement in the interpretation of medical images and reports enables the analysis of diagnostic images, such as chest X-rays (CXRs), through textural prompts which can include indications, reports, and any other text inputs that we want the model to leverage. These models have shown strong performance in unseen tasks, and this versatility and robustness highlight the potential of these models as de-facto models for CAD \cite{li2023comprehensive,gpt4,tu2023towards}.

For CXRs, VLMs have demonstrated utility in several tasks, including the automatic generation of radiology findings from images, visual question answering of these images, and correction of radiology reports based on CXR images \cite{wu2023exploring,wu2024hallucination,microsoftmultimodal2024}. Through these capabilities, VLMs not only streamline the diagnostic workflow but also offer valuable insights that can aid radiologists in making informed decisions. Although these models are designed to serve as interactive assistants, the human-computer interaction is limited to input CXR images and text prompts.

To further enhance human-computer interaction in CXR analysis using VLMs, we propose a novel method of integrating eye gaze data into the VLM framework for CXRs. This approach utilizes heatmaps generated from eye gaze patterns recorded during the interpretation of images. By incorporating these heatmaps, we introduce an additional layer of insight to the VLM, representing an advancement in human-centred AI in medical image analysis. We also further fine-tuned the models with eye gaze data. In this work, we tested our models’ effectiveness in four clinical tasks: Report Automation, Error Detection, Differential Diagnosis, and Visual Question and Answering. 

Our contributions in this paper are as follows:
\begin{enumerate}
    \item \textbf{Enhancing Human-Computer Interaction for VLMs with Eye Gaze}: This work used heatmaps with VLMs for clinical applications for the first time, which highlight the precise focal points and duration of a radiologist's attention when analysing a CXR. 
    \item \textbf{Comprehensive evaluation on 4 real-world clinical applications}: We test the effectiveness of our approach in a comprehensive list of models (10 models including our own fine-tuned model) on 4 clinical applications: Report Automation, Error Detection, Differential Diagnosis and Visual Question and Answering. 
\end{enumerate}

\begin{figure}
    \centering
    \includegraphics[width=1\linewidth]{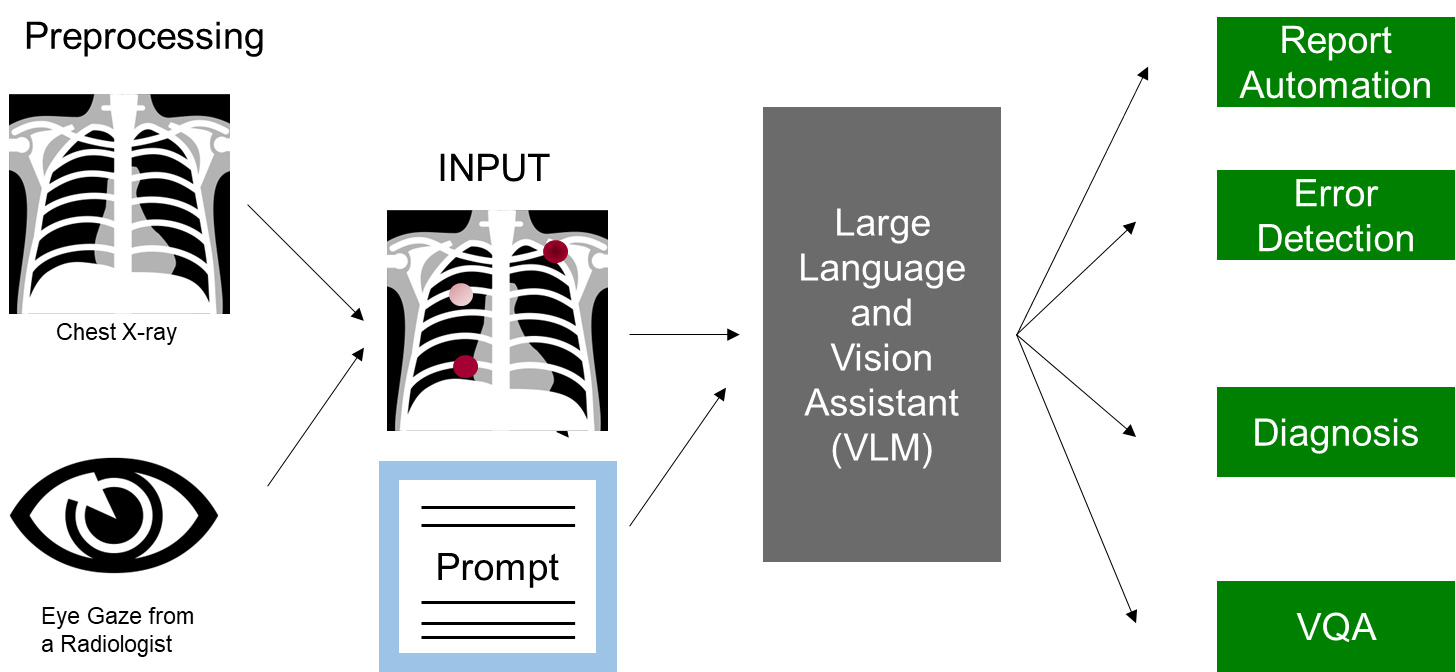}
    \caption{Overview of Enhancing Human-Computer Interaction in Chest X-ray Analysis using Vision and Language Model with Eye Gaze Patterns.} 
    \label{fig:fig1}
\end{figure}

\section{Related Work}
Several studies have shown promising results by incorporating radiologists' eye-tracking data into AI models, leading to improved diagnostic accuracy \cite{ma2023hitleye,ji2023hitlmammo}. For example, the EG-ViT model leverages radiologists' eye gaze data to guide its attention towards potentially pathological regions \cite{ma2023hitleye}. Another model Mammo-Net explores enhancing mammogram classification by incorporating eye gaze data and multi-view information to improve mammogram classification, which addresses interpretability and data annotation limitations by using radiologists' eye gaze data \cite{ji2023hitlmammo}. However, these models typically focus on analyzing only images, which constrains the usage of the model to single-modality applications.

% \subsection{EG-ViT}
% EG-ViT model leverages radiologists' eye gaze data to guide its attention towards potentially pathological regions \cite{ma2023hitleye}. This approach helped the model avoid learning irrelevant patterns and improved the reliability of its results. However, EG-ViT is limited to processing only images.

% \subsection{Mammo-Net}
% Mammo-Net explores enhancing mammogram classification by incorporating eye gaze data and multi-view information to improve mammogram classification \cite{ji2023hitlmammo}. This work addresses interpretability and data annotation limitations by using radiologists' eye gaze data. While Mammo-Net demonstrates a significant improvement in both classification performance and interpretability, it lacks the versatility of multimodal VLMs. 

Textual data, such as clinical notes, are equally critical in the clinical setting, containing detailed and comprehensive information. Thus, we believe embedding both images and reports in Visual Language Models (VLMs) can achieve a more holistic understanding and improve diagnostic accuracy.

% Unlike the aforementioned models, VLMs can process both radiology images and textual data (such as radiology reports) within a unified framework. This allows them to exploit not only the visual information in the images but also the rich context provided by reports.

\section{Tasks}
Figure \ref{fig:fig1} summarizes an overview of our CXR analysis framework utilizing eye gaze patterns in the following 4 tasks in CAD. This figure shows how the input CXR image is processed to incorporate eye gaze patterns. Paired with textual prompts for the downstream task, the eye gaze heatmap-enhanced CXR images provide extra human intelligence to the VLM.

\subsection{Report Automation (GEN and SUM)}

Radiology report automation typically involves generating and summarizing radiology reports through VLMs. A standard radiology report includes a "Findings" section, which describes the observations made from the images, and an "Impressions" section, which offers a summary for diagnostic purposes. The task of report generation, called hereafter as \textbf{GEN}, aims to produce the "Findings" section based on the images. In contrast, report summarization, called hereafter as \textbf{SUM},  seeks to construct the "Impressions" section by leveraging both the "Findings" and the images.

\subsection{Error Detection (ERR)}

In light of errors found in radiologists' reports, ensuring diagnostic accuracy is imperative \cite{brady2017error}. It is shown by other work leveraging VLMs as supportive tools can help assist radiologists in identifying errors that appeared in reports \cite{wu2023exploring}. For such error detection tasks, we created the evaluation data by introducing synthetic errors into the original report to test VLM's ability to classify their presence ("Y") or absence ("N"). Errors were introduced randomly on important phenotypical sentences\cite{wu2023exploring}, and some of the reports were left unchanged.

\subsection{Differential Diagnosis (DDx)}

In this task, the model generates potential diagnoses from chest X-ray images. Given that diagnoses are commonly classified using the International Classification of Diseases (ICD) codes, we refine the model's raw text output to correspond with these classifications. For disease entity recognition within the text, we employ a DeBERTa-V3-large model, fine-tuned on the BC5CDR dataset \cite{he2021debertav3,ushio2022tner,wei2016bc5cdr}. To ensure accurate disease entity alignment with ICD codes, we utilize embeddings from a SapBERT model \cite{liu2020sapbert}.

\subsection{Visual Question Answering (VQA)}

To evaluate the effectiveness of current state-of-the-art VLMs on the reasoning and understanding of clinical knowledge, we used the MIMIC-CXR-VQA dataset for such evaluation \cite{bae2024mimiccxrvqa}. It is an image-based Electronic Health Records (EHR) question-answering dataset that is designed to facilitate joint reasoning across imaging and table modalities in EHR question-answering (QA) systems.

\section{Methods}
\subsection{Datasets}
In this study, we leverage the posterior to anterior (PA) view images from the MIMIC-Eye dataset, a compilation of MIMIC-IV, MIMIC-CXR, REFLACX, and EyeGaze datasets \cite{hsieh2023mimiceye}. This dataset is comprised of 3,689 chest X-ray images for intensive care unit patients. REFLACX and EyeGaze provide eye-tracking data for radiologists, which is an essential component of our work. While the EyeGaze dataset provides a list of diagnoses, the REFLACX dataset does not provide a list of diagnoses, making the REFLACX dataset unfit for the DDx task.

For the evaluation dataset, We further processed the EyeGaze dataset as it can be used for all the tasks in our study. We use the overlap of the EyeGaze dataset with MIMIC-CXR-VQA and reports with synthetic errors for the \textbf{ERR} task, 574 images. The rest of the EyeGaze dataset and REFLACX dataset were combined to form a train dataset, 1,169 images. We ensured there was no overlap between the evaluation and train dataset. A detailed breakdown of the train and evaluation datasets is provided in Table \ref{tab:data_stat}.

For all the datasets, raw eye gaze information at 1000 Hz is used to create a heat map on the CXR image. We drew a red dot at the x and y coordinate of the gaze with the darkness of the dot representing the number of gazes. In other words, darker dots show more time the radiologist spent time on the spot of the CXR image.

\begin{table}
\centering
\caption{Train and Evaluation Dataset Description.}
\label{tab:data_stat}
\begin{tabular}{|l|l|l|}
\hline
\textbf{Statistics} &  \textbf{Train} & \textbf{Evaluation}\\
\hline
Number of CXR Images  & 1,169          & 574        \\
CXR Images with DDx cases          & 420          & 574        \\
Number of Visual Questions           & 1,542          & 574        \\
Number of Reports     & 940           & 574 \\
Reports Error Rates     &    73.09\%       & 73.17\%        \\
Total Number of Instruction Data  & 4,782          & N/A        \\
\hline
\end{tabular}
\end{table}

\subsection{List of Models}

Table \ref{tab:model_description} shows the models we use for comparison. We include LLaVA variants from open domain (LLaVA-v0 \cite{liu2023llava0}, LLaVA-v1.5 \cite{liu2023llava1} and LLaVA-v1.6 \cite{liu2024llavanext}) and medical domain (LLaVA-Med \cite{li2023llavamed} and CXR-LLaVA \cite{lee2023cxr}). All models except LLaVA-v1.5 and LLaVA-1.6 are based on 7B backbone LLM and LLaVA-v1.5 and LLaVA-1.6 with two model sizes: 7B and 13B. LLaVA-1.6 comes with an additional backbone LLM, Mistral-7B.

% llava-v0 \cite{liu2023llava0}
% llava-v1.5 \cite{liu2023llava1}
% llava-v1.6 \cite{}
% llava-med \cite{li2023llavamed}
% cxr-llava \cite{lee2023cxr}
\begin{table}
\centering
\caption{Model Description.}
\label{tab:model_description}
\resizebox{\textwidth}{!}{%
\begin{tabular}{|l|l|l|l|l|l|}
\hline
\textbf{Model} & \textbf{Vision Encoder} & \textbf{Resolution} & \textbf{Backbone LLM} & \textbf{Connector} & \textbf{Train Data}\\
 &  & \textbf{(pixel)} &  &  & \textbf{Size} \\
\hline
LLaVA-v0  &  CLIP ViT-L/14 & 224  & Vicuna-7B-v0  & Projection  &    753K  \\
LLaVA-Med  &  CLIP ViT-L/14 & 224  & Vicuna-7B-v0  & Projection  &   560K    \\
LLaVA-v1.5-7B  &  CLIP ViT-L/14 & 336  & Vicuna-7B-v1.5  & MLP  &  1223K    \\
LLaVA-v1.5-13B  &  CLIP ViT-L/14 & 336  & Vicuna-13B-v1.5  & MLP  & 1223K     \\
LLaVA-v1.6-7B  &  CLIP ViT-L/14 & 672  & Vicuna-7B-v1.5  & MLP &   1318K     \\
LLaVA-v1.6-13B  &  CLIP ViT-L/14 & 672  & Vicuna-13B-v1.5  & MLP  &  1318K    \\
LLaVA-v1.6M  &  CLIP ViT-L/14 & 672 & Mistral-7B  & MLP  &  1318K    \\
CXR-LLaVA  &  CXR-specific & 512 & LLaMA-7B  & Projection  &  527K    \\
  &   CLIP ViT-L/16 & &  &   &      \\
\hline
\end{tabular}
}
\end{table}

\subsection{Model Train}
Fine-tuning code is from LLaVA's official GitHub repository. We kept the LLaVA's default hyperparameter configurations with two adjustments: the train batch size and the number of epochs. We reduced the train batch size to 8 and limited the number of epochs to 1.

Training our model with 3 A5000 GPUs presented memory limitations. To overcome this challenge, we implemented several techniques that significantly reduced memory consumption. These techniques included low-rank adaptation (LoRA), the DeepSpeed zero-redundancy optimizer (ZeRO3), and flash attention \cite{hu2021lora,rasley2020deepspeed,dao2022flashattention}. 

We trained two versions of one of the open domain models, LLaVA-v1.5-7B: one without the raw CXR images and another one with the eye gaze pattern CXR images. We denote this model with `FT' and `FT+G' respectively.

\subsection{Evaluation}
To ensure efficient and consistent evaluations throughout our experiments, we adopted a zero-shot approach for all tasks and used a batch of 1 with a temperature parameter of 0. We opted temperature to be 0 to minimize the randomness of the model's generated text. For each task, we set specific limits on the maximum length of the model's response: \textbf{GEN} - 320, \textbf{SUM} - 128, \textbf{ERR} - 64, \textbf{DDx} - 192, \textbf{VQA} - 64. These limits were chosen based on the expected length of a typical response in each task. This optimization helps the model perform better and ensures our evaluation results are reliable and consistent across different tasks. To test the effectiveness of the eye gaze pattern, we also evaluated the model with the raw CXR images (`No Gaze') and with the CXR images with the eye gaze pattern (`Gaze')

% \subsubsection{Evaluation Metrics}
We use different evaluation metrics for different tasks. For report automation including \textbf{GEN} and \textbf{SUM}, we use the ROUGE score \cite{lin2004rouge}. It measures the overlap of n-grams, word sequences, and sometimes word pairs between the generated summary and the reference summaries. Among various versions of the ROUGE score, we chose to use ROUGE-L because it focuses on the longest common subsequence. We used HuggingFace's \texttt{evaluate}  package to calculate the score.

For the \textbf{ERR} task, we utilize the accuracy score to evaluate this binary classification performance. To extract a valid answer, we apply regular expression and \texttt{thefuzz} package for string matching.

For \textbf{DDx}, the calculation of the F1 score is adapted to accommodate the specificity of the task. Diagnosis predictions are extracted from the model's responses, focusing on the accuracy at the ICD code level through disease entity recognition and alignment. Precision is computed by dividing the count of correct predictions by the total number of predictions made, while recall is determined by the ratio of correct predictions to the total number of relevant diseases for the patient. These values of recall and precision are then employed to compute the F1 score. For the \textbf{VQA} task, we employ the accuracy score as the metric to assess performance. 

\section{Results and Discussion}

\begin{table}
\centering
\caption{Evaluation Results. \textbf{GEN}: Report Genetaion. \textbf{SUM}: Report Summarization. \textbf{ERR}: Error Detection. \textbf{DDx}: Differential Diagnosis. \textbf{VQA}: Visual Question Answering. The evaluation metrics used in each task are noted in parentheses. No G and G stands for `No Gaze' and `Gaze.' We bold the scores in `Gaze' if they are higher than the corresponding `No Gaze' scores.} 
\label{tab:main_result}
\resizebox{\textwidth}{!}{%
% \usepackage{multirow}
\begin{tabular}{|l|l|ll|ll|ll|ll|ll|}
\hline
\multirow{2}{*}{\textbf{Model}} & \multirow{2}{*}{\textbf{Size}} & \multicolumn{2}{l|}{\textbf{GEN} (R-L)}             & \multicolumn{2}{l|}{\textbf{SUM} (R-L)}             & \multicolumn{2}{l|}{\textbf{ERR} (Acc)}             & \multicolumn{2}{l|}{\textbf{DDx} (F1)}             & \multicolumn{2}{l|}{\textbf{VQA} (Acc)}             \\ \cline{3-12} 
                       &                       & \multicolumn{1}{l|}{\textbf{No G}} & \textbf{G}  & \multicolumn{1}{l|}{\textbf{No G}} & \textbf{G}  & \multicolumn{1}{l|}{\textbf{No G}} & \textbf{G}  & \multicolumn{1}{l|}{\textbf{No G}} & \textbf{G}  & \multicolumn{1}{l|}{\textbf{No G}} & \textbf{G}  \\ \hline
LLaVA-v0               & 7B                    & \multicolumn{1}{l|}{9.86}    & 8.72  & \multicolumn{1}{l|}{9.12}    & 8.89  & \multicolumn{1}{l|}{28.75}   & \textbf{71.78} & \multicolumn{1}{l|}{2.70}    & \textbf{4.59}  & \multicolumn{1}{l|}{43.03}   & 40.07 \\ 
LLaVA-Med              & 7B                    & \multicolumn{1}{l|}{11.39}   & \textbf{12.41} & \multicolumn{1}{l|}{9.99}   & 9.64 & \multicolumn{1}{l|}{70.21}   & \textbf{73.00} & \multicolumn{1}{l|}{2.59}    & \textbf{11.10} & \multicolumn{1}{l|}{46.69}   & 43.03 \\ 
LLaVA-v1.5             & 7B                    & \multicolumn{1}{l|}{15.10}   & 13.56 & \multicolumn{1}{l|}{10.68}   & 9.90 & \multicolumn{1}{l|}{49.13}   & 37.80 & \multicolumn{1}{l|}{2.62}    & \textbf{6.15}  & \multicolumn{1}{l|}{47.74}   & 44.77 \\ 
CXR-LLaVA              & 7B                    & \multicolumn{1}{l|}{24.88}   & 24.60 & \multicolumn{1}{l|}{39.25}   & \textbf{41.43} & \multicolumn{1}{l|}{48.26}   & \textbf{48.61} & \multicolumn{1}{l|}{12.35}   & \textbf{13.31} & \multicolumn{1}{l|}{56.79}   & \textbf{59.06} \\ 
LLaVA-v1.6             & 7B                    & \multicolumn{1}{l|}{11.27}   & 10.04 & \multicolumn{1}{l|}{10.09}   & 10.00 & \multicolumn{1}{l|}{56.27}   & 49.13 & \multicolumn{1}{l|}{7.41}    & \textbf{10.48} & \multicolumn{1}{l|}{44.77}   & \textbf{45.30} \\ 
LLaVA-v1.6M            & 7B                    & \multicolumn{1}{l|}{12.52}   & 11.53 & \multicolumn{1}{l|}{10.23}   & 9.82  & \multicolumn{1}{l|}{58.89}   & 44.95 & \multicolumn{1}{l|}{4.33}    & \textbf{7.05}  & \multicolumn{1}{l|}{56.45}   & \textbf{58.54} \\ 
LLaVA-v1.5             & 13B                   & \multicolumn{1}{l|}{14.97}        &   11.44    & \multicolumn{1}{l|}{11.19}        &     10.46  & \multicolumn{1}{l|}{35.54}        &   28.92    & \multicolumn{1}{l|}{4.25}        &   \textbf{5.64}    & \multicolumn{1}{l|}{45.30}        &   \textbf{46.34}    \\ 
LLaVA-v1.6             & 13B                   & \multicolumn{1}{l|}{11.86}        &  11.41     & \multicolumn{1}{l|}{10.33}        &   10.32    & \multicolumn{1}{l|}{67.42}        &   59.23    & \multicolumn{1}{l|}{5.25}        &   \textbf{6.53}    & \multicolumn{1}{l|}{42.16}        &   42.16    \\ \hline
LLaVA-v1.5FT              &   7B                    & \multicolumn{1}{l|}{29.52}        &   29.30    & \multicolumn{1}{l|}{53.93}        &   53.89    & \multicolumn{1}{l|}{73.87}        &   73.87    & \multicolumn{1}{l|}{23.69}        &   23.69    & \multicolumn{1}{l|}{61.67}        &   61.67    \\
LLaVA-v1.5FT+G              &  7B                     & \multicolumn{1}{l|}{29.32}        &   \textbf{29.76}    & \multicolumn{1}{l|}{53.19}        &    \textbf{53.25}   & \multicolumn{1}{l|}{74.39}        & 74.39      & \multicolumn{1}{l|}{18.16}        &     18.16  & \multicolumn{1}{l|}{52.79}        &  \textbf{57.49}     \\
\hline
\end{tabular}
}
\end{table}

Analysing the inference performance of VLMs with eye gaze patterns with those without highlighted intriguing insights in these models as well as the incorporation of this extra human intelligence. The comparison evaluation results are detailed in Table \ref{tab:main_result}. 

\subsection{Eye gaze enhances DDx most evidently}
All the baseline models perform better with eye gaze patterns in DDx. The largest increase from `No Gaze’ (2.59\%) to `Gaze’ (11.10\%) was seen by the \textbf{LLaVA-Med model}. This increase in performance is not seen in all the models in other tasks. Still, we observe the improvement in at least one baseline model in all the tasks. In the error detection task, we see the largest increase in \textbf{LLaVA-v0}, a 43.03\% increase from `No Gaze’ (28.75\%) to `Gaze’ (71.78\%). \textbf{LLaVA-Med} and \textbf{CXR-LLaVA} performance also increased with eye gaze patterns. Apart from these models, \textbf{LLaVA-v1.6}, \textbf{LLaVA-v1.6M}, and \textbf{LLaVA-v1.5-13B} models saw an increase in performance for the \textbf{VQA} task. These findings highlight the effectiveness of incorporating this additional human-computer interaction to enhance model performance.

\subsection{Medical models perform better with eye gaze patterns}
The evaluation result also highlighted significant performance enhancements in models fine-tuned within the medical domain data, particularly \textbf{LLaVA-Med} and \textbf{CXR-LLaVA}. \textbf{LLaVA-Med}, which was fine-tuned with PMC figure and legends, show an increase in performance with eye gaze pattern for \textbf{GEN}, \textbf{ERR}, and \textbf{DDx}. \textbf{CXR-LLaVA} model, which was fine-tuned with MIMIC reports, saw an increase in performance with the `Gaze’ image in all tasks except \textbf{GEN}. The result suggests that domain-specific fine-tuning can significantly enhance VLM performance in clinical applications and may play a role in this improvement of performance with eye gaze patterns.

\subsection{Larger models do not perform well}
Interestingly, our evaluation revealed that larger models did not consistently perform better in some tasks. The LLaVA-v1.5-13B model performed worse than its 7B model in \textbf{GEN}, \textbf{ERR}, and \textbf{VQA}.
Also, this trend is observed in LLaVA-v1.6 models: 7B model outperforming in \textbf{DDx} and \textbf{VQA}. This result suggests that adding more parameters does not directly translate to performance improvements.

\subsection{Impact of eye gaze patterns in fine-tuning}
We selected the LLaVA-v1.5-7B model because it was the best-performing model apart from the LLaVA-v1.6 models, for which the training code was not publicly available. Both versions of LLaVA-v1.5-7B fine-tuned models exhibited superior performance over all the other models, including the medical models, except for the \textbf{VQA} task. Also, the eye gaze pattern proved to be promising in enhancing the model's performance for \textbf{GEN} and \textbf{ERR} tasks which the baseline model actually showed a decrease in performance. 

These findings pave the way for further exploration of eye gaze data integration in VLM-based clinical applications. By leveraging radiologists' eye gaze data, we showed promise to enhance VLMs for more accurate and insightful clinical decision support systems in medical imaging analysis. Future research could delve deeper into the interplay between eye gaze patterns and model architectures to unlock the full potential of these collaborative AI systems in healthcare. Also, in the future, this work can be extended to other types of medical datasets such as CT and MRI as well as other tasks in CAD such as segmentation.

\section{Conclusion}

In conclusion, our study has demonstrated the potential of integrating radiologists’ eye-tracking data into VLMs to enhance the accuracy of analyzing chest X-rays. This is particularly evident in the area of differential diagnosis, where all the baseline models we tested achieved a remarkable improvement with eye gaze patterns. Also, the positive impact of eye gaze data on fine-tuned models suggests its potential to enhance performance even in tasks where the baseline model struggles. This enhanced human-computer interaction approach promises to improve decision-making in clinical practices, suggesting a pivotal step toward a more synergistic human-AI collaboration.

\bibliographystyle{splncs04}
\bibliography{miccai}

\begin{thebibliography}{10}
\providecommand{\url}[1]{\texttt{#1}}
\providecommand{\urlprefix}{URL }
\providecommand{\doi}[1]{https://doi.org/#1}

\bibitem{bae2024mimiccxrvqa}
Bae, S., Kyung, D., Ryu, J., Cho, E., Lee, G., Kweon, S., Oh, J., Ji, L., Chang, E., Kim, T., et~al.: Ehrxqa: A multi-modal question answering dataset for electronic health records with chest x-ray images. Advances in Neural Information Processing Systems  \textbf{36} (2024)

\bibitem{brady2017error}
Brady, A.P.: Error and discrepancy in radiology: inevitable or avoidable? Insights into imaging  \textbf{8},  171--182 (2017)

\bibitem{calisto2022hitlbreast}
Calisto, F.M., Santiago, C., Nunes, N., Nascimento, J.C.: Breastscreening-ai: Evaluating medical intelligent agents for human-ai interactions. Artificial Intelligence in Medicine  \textbf{127},  102285 (2022)

\bibitem{dao2022flashattention}
Dao, T., Fu, D., Ermon, S., Rudra, A., R{\'e}, C.: Flashattention: Fast and memory-efficient exact attention with io-awareness. Advances in Neural Information Processing Systems  \textbf{35},  16344--16359 (2022)

\bibitem{he2021debertav3}
He, P., Gao, J., Chen, W.: Debertav3: Improving deberta using electra-style pre-training with gradient-disentangled embedding sharing. arXiv preprint arXiv:2111.09543  (2021)

\bibitem{hsieh2023mimiceye}
Hsieh, C., Ouyang, C., Nascimento, J.C., Pereira, J., Jorge, J., Moreira, C.: Mimic-eye: Integrating mimic datasets with reflacx and eye gaze for multimodal deep learning applications  (2023)

\bibitem{hu2021lora}
Hu, E.J., Shen, Y., Wallis, P., Allen-Zhu, Z., Li, Y., Wang, S., Wang, L., Chen, W.: Lora: Low-rank adaptation of large language models. arXiv preprint arXiv:2106.09685  (2021)

\bibitem{hwang2021deep}
Hwang, E.J., Lee, J.H., Kim, J.H., Lim, W.H., Goo, J.M., Park, C.M.: Deep learning computer-aided detection system for pneumonia in febrile neutropenia patients: a diagnostic cohort study. BMC Pulmonary Medicine  \textbf{21}(1), ~406 (2021). \doi{10.1186/s12890-021-01768-0}, \url{https://doi.org/10.1186/s12890-021-01768-0}

\bibitem{ji2023hitlmammo}
Ji, C., Du, C., Zhang, Q., Wang, S., Ma, C., Xie, J., Zhou, Y., He, H., Shen, D.: Mammo-net: Integrating gaze supervision and interactive information in multi-view mammogram classification. In: International Conference on Medical Image Computing and Computer-Assisted Intervention. pp. 68--78. Springer (2023)

\bibitem{lee2023cxr}
Lee, S., Youn, J., Kim, M., Yoon, S.H.: Cxr-llava: Multimodal large language model for interpreting chest x-ray images. arXiv preprint arXiv:2310.18341  (2023)

\bibitem{li2023llavamed}
Li, C., Wong, C., Zhang, S., Usuyama, N., Liu, H., Yang, J., Naumann, T., Poon, H., Gao, J.: Llava-med: Training a large language-and-vision assistant for biomedicine in one day. arXiv preprint arXiv:2306.00890  (2023)

\bibitem{li2023comprehensive}
Li, Y., Liu, Y., Wang, Z., Liang, X., Liu, L., Wang, L., Cui, L., Tu, Z., Wang, L., Zhou, L.: A comprehensive study of gpt-4v's multimodal capabilities in medical imaging. medRxiv pp. 2023--11 (2023)

\bibitem{lin2004rouge}
Lin, C.Y.: Rouge: A package for automatic evaluation of summaries. In: Text summarization branches out. pp. 74--81 (2004)

\bibitem{liu2020sapbert}
Liu, F., Shareghi, E., Meng, Z., Basaldella, M., Collier, N.: Self-alignment pretraining for biomedical entity representations. arXiv preprint arXiv:2010.11784  (2020)

\bibitem{liu2023llava1}
Liu, H., Li, C., Li, Y., Lee, Y.J.: Improved baselines with visual instruction tuning. arXiv preprint arXiv:2310.03744  (2023)

\bibitem{liu2024llavanext}
Liu, H., Li, C., Li, Y., Li, B., Zhang, Y., Shen, S., Lee, Y.J.: Llava-next: Improved reasoning, ocr, and world knowledge (January 2024), \url{https://llava-vl.github.io/blog/2024-01-30-llava-next/}

\bibitem{liu2023llava0}
Liu, H., Li, C., Wu, Q., Lee, Y.J.: Visual instruction tuning. arXiv preprint arXiv:2304.08485  (2023)

\bibitem{ma2023hitleye}
Ma, C., Zhao, L., Chen, Y., Wang, S., Guo, L., Zhang, T., Shen, D., Jiang, X., Liu, T.: Eye-gaze-guided vision transformer for rectifying shortcut learning. IEEE Transactions on Medical Imaging  (2023)

\bibitem{gpt4}
OpenAI: Gpt-4 (2023), \url{https://www.openai.com/gpt-4}

\bibitem{hitlperformance}
Patel, B.N., Rosenberg, L., Willcox, G., Baltaxe, D., Lyons, M., Irvin, J., Rajpurkar, P., Amrhein, T., Gupta, R., Halabi, S., Langlotz, C., Lo, E., Mammarappallil, J., Mariano, A.J., Riley, G., Seekins, J., Shen, L., Zucker, E., Lungren, M.P.: Human--machine partnership with artificial intelligence for chest radiograph diagnosis. npj Digital Medicine  \textbf{2}(1), ~111 (2019). \doi{10.1038/s41746-019-0189-7}, \url{https://doi.org/10.1038/s41746-019-0189-7}

\bibitem{cadradiographyreview}
Qin, C., Yao, D., Shi, Y., Song, Z.: Computer-aided detection in chest radiography based on artificial intelligence: a survey. BioMedical Engineering OnLine  \textbf{17}(1), ~113 (2018). \doi{10.1186/s12938-018-0544-y}, \url{https://doi.org/10.1186/s12938-018-0544-y}

\bibitem{rasley2020deepspeed}
Rasley, J., Rajbhandari, S., Ruwase, O., He, Y.: Deepspeed: System optimizations enable training deep learning models with over 100 billion parameters. In: Proceedings of the 26th ACM SIGKDD International Conference on Knowledge Discovery \& Data Mining. pp. 3505--3506 (2020)

\bibitem{cadCOVIDimprovement}
Shaheed, K., Szczuko, P., Abbas, Q., Hussain, A., Albathan, M.: Computer-aided diagnosis of covid-19 from chest x-ray images using hybrid-features and random forest classifier. Healthcare  \textbf{11}(6) (2023). \doi{10.3390/healthcare11060837}, \url{https://www.mdpi.com/2227-9032/11/6/837}

\bibitem{tu2023towards}
Tu, T., Azizi, S., Driess, D., Schaekermann, M., Amin, M., Chang, P.C., Carroll, A., Lau, C., Tanno, R., Ktena, I., et~al.: Towards generalist biomedical ai. arXiv preprint arXiv:2307.14334  (2023)

\bibitem{ushio2022tner}
Ushio, A., Camacho-Collados, J.: T-ner: an all-round python library for transformer-based named entity recognition. arXiv preprint arXiv:2209.12616  (2022)

\bibitem{wei2016bc5cdr}
Wei, C.H., Peng, Y., Leaman, R., Davis, A.P., Mattingly, C.J., Li, J., Wiegers, T.C., Lu, Z.: Assessing the state of the art in biomedical relation extraction: overview of the biocreative v chemical-disease relation (cdr) task. Database  \textbf{2016} (2016)

\bibitem{wu2023exploring}
Wu, J., Kim, Y., Keller, E.C., Chow, J., Levine, A.P., Pontikos, N., Ibrahim, Z., Taylor, P., Williams, M.C., Wu, H.: Exploring multimodal large language models for radiology report error-checking. arXiv preprint arXiv:2312.13103  (2023)

\bibitem{wu2024hallucination}
Wu, J., Kim, Y., Wu, H.: Hallucination benchmark in medical visual question answering. arXiv preprint arXiv:2401.05827  (2024)

\bibitem{microsoftmultimodal2024}
Yildirim, N., Richardson, H., Wetscherek, M.T., Bajwa, J., Jacob, J., Pinnock, M.A., Harris, S., de~Castro, D.C., Bannur, S., Hyland, S.L., et~al.: Multimodal healthcare ai: Identifying and designing clinically relevant vision-language applications for radiology. arXiv preprint arXiv:2402.14252  (2024)

\end{thebibliography}




% LNCS Homepage, \url{http://www.springer.com/lncs}, last accessed 2023/10/25
% \end{thebibliography}
\end{document}